%% file: lrec2026-example.tex
\documentclass[10pt, a4paper]{article}

\usepackage[final]{lrec2026} %
\usepackage{graphicx}
\usepackage{tabularx}
\usepackage{soul}
\usepackage{rotating}
\usepackage{comment}
\usepackage{multirow}
\usepackage{natbib}

\usepackage{url}

\usepackage{xcolor}
\definecolor{darkgreen}{rgb}{0.0,0.5,0.0}\usepackage{booktabs}

\definecolor{sigdif}{rgb}{0.9, 0.9, 0.9}
\definecolor{NA}{rgb}{0.9, 0.9, 0.9}

\input{macros_Jirka.tex}

\setlingexlabelmax{99}

\usepackage{alltt}

\usepackage{enumitem}
\setlist[itemize]{topsep=5pt}
\setlist[enumerate]{topsep=5pt}
\setlist{itemsep=3pt} %

\title{Introducing corpora Hlava Cor and Hlava AD: Human Label Variation in Coreference and Discourse Relations}
\name{Anna Nedoluzhko, Šárka Zikánová, Jiří Mírovský, Milan Straka and Eva Hajičová} 

\address{Charles University, Faculty of Mathematics and Physics, Institute of Formal and Applied Linguistics \\
         Prague, Czech Republic \\
         \{nedoluzhko, zikanova, mirovsky, straka, hajicova\}@ufal.mff.cuni.cz\\}

\abstract{
As previous research on annotator disagreement in discourse phenomena has shown, understanding text coherence varies considerably from one individual to another. To explore this phenomenon, we created two corpora with multiple annotations of Czech texts, accompanied by annotators’ explanations of their choices. The first corpus consists of 1,024 contexts annotated in parallel by three annotators. It captures differences in the identification of coreference across various text types and grammatical-semantic categories, including pronouns, full noun phrases, and anaphoric adverbials. The second corpus comprises 512 contexts, annotated in parallel by five annotators, and focuses on identifying discourse relations in attributive and non-attributive constructions. Both corpora achieve a comparable inter-annotator agreement of approximately 60–65\%. For coreference annotation, agreement tends to be lower in cases where automatic coreference resolution models disagree, suggesting that when the models disagree, the examples tend to be more difficult or ambiguous for human annotators to interpret. The annotators’ comments, both for coreference and discourse relations, further reveal differences in interpretation, varying levels of confidence in text understanding, and individual reading strategies.\\ \newline  }

\usepackage{fancyhdr}
\fancypagestyle{officialbibref}{\fancyhf{}\fancyheadoffset[L,R]{50pt}\fancyhead[C]{This paper was presented at \textbf{SLiDE 2026} -- please cite that version instead.}\fancyfoot[C]{\normalsize\thepage}\addtolength{\topmargin}{-50pt}\addtolength{\headsep}{50pt}}
\begin{document}
\thispagestyle{officialbibref}
\pagenumbering{arabic}\pagestyle{plain}

\maketitleabstract

\enlargethispage{\baselineskip}

\section{Introduction}

During our long-term development of corpora focused on discourse relations, coreference, and information structure, we observed that certain linguistic phenomena tend to produce lower inter-annotator agreement than others. This observation aligns with a growing body of research on annotation evaluation taking into account different perspectives (\citealp{Basile2021Disagreement}) and Human Label Variation (HLV, \citealp{Plank2022Problem}), which suggests that disagreement in NLP is often a reflection of inherent linguistic complexity rather than mere noise.
It is possible to identify several factors that appear to contribute to this variation, ranging from the polysemy of synsemantic signals to word order patterns (see, e.g., \citealp{biblio:ZiTextStructure2024}). Furthermore, annotator-related factors, such as the subjective perception of the relative importance of text segments, play a significant role. Such interpretative diversity is well-documented in large-scale projects; for instance, Poesio et al. \citeyearpar{Poesio2020Ambiguity} and Recasens et al. \citeyearpar{Recasens2010AnCora} have consistently observed that ambiguity in textual meaning naturally reduces agreement in coreference tasks.

In the present paper, we introduce two corpora created to investigate human label variation. \textbf{Hlava Cor} (Human Label Variation in Coreference \citelanguageresource{hlavacor}) contains annotations of coreference and focuses on human label variation related to genericity, subjectivity and underspecification. 
\textbf{Hlava AD} (Human Label Variation in Attribution and Discourse \citelanguageresource{hlavaad}) comprises annotations of discourse relations in attributive and non-attributive constructions.

In developing these corpora, we took into account various hypotheses about the possible causes of annotation disagreement, based on our previous annotation experience and analyses.
For the coreference corpus, we specifically distinguish generic and specific expressions, as well as different grammatical–semantic categories, including pronouns, full noun phrases, and anaphoric adverbials.
We further hypothesize that coreference resolution models may partially predict cases of underspecification, in the sense that where models disagree, human annotators are also likely to find the interpretation of coreference relations problematic (see Section~\ref{statistics} and \ref{Hlava CorIAA}).

For discourse relations, we assume that in texts with larger segments of direct or indirect speech, it may be more difficult for recipients to recognize which of the upcoming sentences are still related to the direct/indirect speech (see Section~\ref{Hlava AD}).

We argue that multiple annotation, and especially the detailed comments accompanying each annotation, provide rich material for investigating the reasons and nuances of human label variation, enhancing our understanding of how people interpret texts differently. We understand our corpora as datasets providing many types of commented structures, and thus serving as a base for future psycholinguistic experiments dealing with special cases. 
Furthermore, these corpora may serve as reference datasets for evaluating the reliability of previously single-annotated corpora.

\section{Related Work}
The interpretation of linguistic meaning is never entirely fixed or uniform; it is shaped by context, perspective, and the inherent ambiguity of natural language. This fundamental indeterminacy influences all levels of linguistic analysis, including discourse and coreference annotation, where human judgments often diverge even under detailed annotation guidelines. %
Researchers in discourse annotation address this challenge in different ways. For example, Marchal et al.~\citeyearpar{Marchal2022Establishing} discuss methods evaluating annotation quality across multiple annotators. 
Building on the "ecologically valid" explanations introduced by Jiang et al.~\citeyearpar{jiang-etal-2023-ecologically}, recent work by Weber-Genzel et al.~\citeyearpar{weber-genzel-etal-2024-varierr} utilizes a two-round annotation procedure where participants justify their label choices through natural language explanations. In these studies, annotators classify the relations between clauses as entailment, contradiction, or neutral and provide commentaries that allows researchers to distinguish between interpretative variation and annotation errors.
Other studies emphasize the importance of explicitly capturing interpretative plurality (\citealp{Plank2022Problem}, \citealp{Basile2021Disagreement}, \citealp{Crible2019Functions}).

New corpora are published capturing multiple understandings of discourse relations, e.g. in the RST approach (\citealp{peng-etal-2022-gcdt},  \citealp{polakova-etal-2024-developing}, \citealp{hewett-stede-2025-disagreements}), or implicit discourse relations in the PDTB approach (\citealp{scholman-etal-2022-discogem}, \citealp{yung-etal-2024-discogem}). None of these datasets include annotators' comments on their choices.

Comparative research on spoken and written data further demonstrates that modality contributes to interpretative diversity: spoken language tends to involve more implicit and multifunctional relations (\citealp{rehbein2016Annotating}, \citealp{Cuenca2017DiscourseMarkers}, \citealp{Crible2019Functions}), though its annotation consistency is not necessarily lower than that of written texts~\cite{Zufferey2015Unified}.

Ambiguity in textual meaning naturally reduces inter-annotator agreement (IAA), as has been consistently observed in large-scale coreference annotation projects (\citealp{Weischedel2011Ontonotes}; \citealp{Zeldes2017GUM}; \citealp{Recasens2010AnCora}; \citealp{Poesio2020Ambiguity}). While analyses of annotation disagreement have occasionally been included (\citealp{Recasens2011Identity}; \citealp{Pradhan2012CoNLL}%
), such studies are typically limited to small, task-specific subsets of data (cf. \citealp{levine-zeldes-2025-subjectivity}). A few corpora explicitly encode ambiguous or near-identity cases (\citealp{Uryupina2020ARRAU}; \citealp{Bourgonje2020Potsdam}; \citealp{Ogrodniczuk2013Polish}), but these remain statistically rare, leaving much of the natural variation in human interpretation unexplored. 

Poesio et al.~\citeyearpar{poesio-etal-2019-crowdsourced} present a preliminary analysis of disagreements in a corpus of anaphoric information in 542 English documents crowdsourced through a game-with-a-purpose. The corpus contains multiple concurrent annotations of about 108 thousand markables, with 20 judgments on average per markable (12 annotations and 8 validations). Expert analysis of a sample of the data shows, however, that genuine ambiguity occurs only in about 9\% of the markables and the other cases of annotators’ disagreement should be accounted to annotators’ errors and to various limitations of the coding scheme and the annotation interface. 

In our analysis, we use the Prague Dependency Treebank - Consolidated 2.0 (PDT-C 2.0; \citelanguageresource{pdtc20}) as the primary data source for extracting and then multiply annotating the data.%

Namely, we use its subcorpus Prague Dependency Treebank (PDT) for the written mode and Prague Dependency Treebank of Spoken Czech (PDTSC) for the spoken mode. The PDT represents texts from the Czech newspapers from 1990's, while the PDTSC contains transcripts of conversations between interviewers and Holocaust survivors or contemporary witnesses of communism. 
The PDT-C~2.0 corpus contains manual annotations of coreference and discourse relations, along with multiple linguistic layers ranging from morphology to tectogrammatics. Annotation in PDT-C 2.0 follows the principle of a single correct decision, with approximately 10\% of the data annotated in parallel to assess inter-annotator agreement (IAA). The collection comprises both spoken and written subcorpora, enabling comparative analyses of discourse phenomena across modalities.

\section{Hlava Cor and Hlava AD: General Settings}

To study human label variation, we extracted texts primarily from different parts of the PDT-C~2.0 (see detailed description in Sections~\ref{statistics} and \ref{Hlava AD}) and created Hlava Cor and Hlava AD. The overall statistics for the annotated data are listed in Table~\ref{tab:corpus_stats}. Hlava Cor contains 1,024 cases annotated by three annotators, while Hlava AD contains 512 cases annotated by five annotators.\footnote{The discrepancy in the size of the two corpora and the number of participating annotators does not stem from a specific theoretical or methodological requirement, but rather reflects the successive stages of their development and the varying availability of resources during each phase of the project.}

\begin{table}[h]
\caption{Overall Statistics for Hlava Cor and Hlava AD}
\label{tab:corpus_stats}
\resizebox{\columnwidth}{!}{%
\centering
\begin{tabular}{lcc}
\toprule
\textbf{Corpus Name} & \textbf{Number of Cases} & \textbf{Number of Annotators} \\
\midrule
Hlava Cor & 1,024 & 3 \\
Hlava AD & ~~512 & 5 \\
\bottomrule
\end{tabular}
}
\end{table}

Both corpora consist of short Czech text segments containing the relevant linguistic phenomena. 
Each text segment was multiply annotated by several annotators. The annotations were performed on the linear texts presented in an Excel spreadsheet.

The annotation team consisted of native Czech speakers, primarily philology students (aged 18–30) with a basic linguistic background but no formal training in specific theoretical frameworks. This lack of prior exposure was intentional to ensure that text comprehension was not biased by theoretical presuppositions. While a core group of five annotators worked on Hlava AD, the Hlava Cor corpus was annotated by six individuals organized into two groups by three annotators. This larger group included the same five participants from the Hlava AD task plus one additional annotator.

The annotation process is designed not only to capture the final annotation decisions but also to elicit the annotators’ interpretative considerations.
For this reason, all annotations obligatorily include annotators' comments explaining their choices %
which are crucial for our investigation.

\section{Hlava Cor: Human Language Variation in Coreference}
\label{Hlava Cor}
The annotation task for Hlava Cor was defined as the identification of coreference, i.e. reference to the same extra-linguistic entity, concept, or situation.

\subsection{Hlava Cor: Annotation Process}
\label{Hlava Coranotproces}
Annotators were instructed to read the column \texttt{Sentence} with a highlighted expression in question and the left-hand context in columns \texttt{Adjacent Context} and \texttt{Distant Context} (see Table~\ref{tab:priklad_anotace})\footnote{The column \texttt{Distant Context} is not shown in the example for space reasons.} and to find a potential antecedent in the preceding context if any. No potential antecedents have been pre-annotated.\footnote{In Table~\ref{tab:priklad_anotace}, two antecedents are highlighted in bold because the table shows an already completed annotation. The bold highlighting is used only to make the example easier to read in this paper; the annotators did not see these highlights during the annotation process.}
Each annotation output contains (i) the antecedent expression if it appeared in the preceding context, (ii) a numeric value indicating the degree to which context was required for interpretation (0 = no distant context needed, 1 = distant context necessary, 2 = even wider context needed), and (iii) a free-text comment explaining the annotator’s reasoning or uncertainty.

The annotation guidelines specified detailed conventions for the form and syntactic scope of antecedents. Annotators were required to record the full nominal phrase as it appears in the text, excluding prepositions unless they constitute part of a named entity or are internal to the referring expression. Dependent modifiers were to be included, while relative clauses were to be excluded. Deviations from these rules had to be explicitly documented in the comment field. Further specifications addressed the treatment of references to clauses or larger textual segments, possessive and other adjectival expressions, coordinated and discontinuous antecedents. When no suitable antecedent was found, annotators entered NE (“no antecedent”), and in cases of complete indecision they recorded a question mark.

\subsection{Hlava Cor: Annotation Example}
An example of the annotated segment is presented in Table~\ref{tab:priklad_anotace}. For the expression \textit{vozidlo} ‘vehicle’ in the column \texttt{Sentence}, the annotators were asked to identify an antecedent (if any) and record it in the \texttt{ANN\_ante} column. In the \texttt{ANN\_comment} column, the annotators were asked to justify their decision or provide any additional remarks.

This example illustrates the annotation by two annotators out of three who arrived at different conclusions. Annotator~1 (ANN1) selected \textit{vůz} ‘car’ as the antecedent of \textit{vozidlu} ‘vehicle’ in \texttt{Sentence}, arguing that the expression does not refer to the specifically described model driven by the journalist, but rather to the vehicle in general. Annotator~2 (ANN2), on the other hand, chose \textit{vozu} ‘car (genitive)’ as the antecedent, noting that both expressions refer to the same entity.

For space reasons, the annotation of the third annotator is not shown; this annotator made the same decision as ANN1 but expressed more uncertainty in the comment. We also omit the annotation of the need for context, which was zero for all three annotators in this case. %

\begin{table*}[h!]
\caption{Example of coreference annotation by two annotators}
\label{tab:priklad_anotace}
\resizebox{\textwidth}{!}{%
\scriptsize %
\setlength{\tabcolsep}{1.5pt} %
\begin{tabular}{|p{4.5cm}|p{3cm}|c|p{3.5cm}|c|p{3.5cm}|}
\hline
\textbf{Adjacent Context} & \textbf{Sentence} & \textbf{ANN1 ante} & \textbf{ANN1 comment} & \textbf{ANN2 ante} & \textbf{ANN2 comment} \\
\hline
Firma Renault Česká republika nám k redakčnímu testu poskytla verzi s turbodieselovým motorem s označením RT. Sice jsme v úvodu uvedli, komu lze \textbf{vůz} [ANN1] především doporučit. Snad nám nebudou mít případní zájemci z kategorie `mladá rodina s více dětmi' za zlé, když jim neprozradíme, kde vzít potřebných 969820 korun. V případě testovaného \textbf{vozu} [ANN2] dovybaveného dvěma střešními okny, autorádiem s ovládáním pod volantem a ve vlnové metalíze pak ještě přes 66 tisíc korun navrch. & Věnujeme se však raději samotnému \textbf{\textcolor{blue}{vozidlu}}, neboť svezení s ním opravdu stojí za to. & vůz & Neodkazuje na detailně popsaný, konkrétní model vozu, s nímž zrovna jel novinář, ale obecně na vozidlo. & vozu & Oba výrazy označují stejnou skutečnost. \\
\hline
\end{tabular}
}
\begin{flushleft}
\vspace{1mm}
\footnotesize %
\textbf{\underline{Adjacent Context}:} The company Renault Czech Republic provided us with a version featuring a turbo diesel engine, designated RT, for an editorial test. Although we mentioned at the beginning who this \textbf{car} [ANN1] can primarily be recommended to, we hope that potential buyers from the category of “young families with several children” will not hold it against us if we do not reveal where to get the required 969,820 crowns. In the case of the tested \textbf{car} [ANN2], additionally equipped with two sunroofs, a car radio with steering-wheel controls, and metallic paint, the price increases by more than 66,000 crowns.
\vspace{1mm}\\
\textbf{\underline{Sentence}:} However, let us turn to the \textbf{\textcolor{blue}{vehicle}} itself, since driving it is really worth the experience.\\
\textbf{\underline{ANN1 ante:}} \textit{vůz} ‘car' \\
\textbf{\underline{ANN1 comment:}} Does not refer to the specifically described model that the journalist was driving, but rather to the vehicle in general.\\
\textbf{\underline{ANN2 ante:}} \textit{vozu} ‘car'(genitive)\\
\textbf{\underline{ANN2 comment:}} Both expressions refer to the same entity.
\end{flushleft}
\end{table*}

\subsection{Hlava Cor: Data Description and Statistics}
\label{statistics}
The corpus consists of 1,024 cases, each marked by three annotators. 
The segments for annotation were selected according to several dimensions of data division. First, we took the same number of spoken and written text segments. 
Most of the text segments have been excerpted from the PDT-C~2.0, a few segments come from the Czech Academic Corpus 2.0 (CAC) ~\citelanguageresource{cac2}. For the written data, we additionally used examples from the archive %
of iRozhlas.\footnote{the internet archive of the Czech news outlet iRozhlas, \url{https://www.irozhlas.cz/zpravy-archiv}}
Table~\ref{tab_cor_spoken} shows the data distribution from these resources. %

\begin{table}[h]
\caption{Structure of Hlava Cor: written vs. spoken modes}
\label{tab_cor_spoken}
\resizebox{\columnwidth}{!}{%
\centering
\begin{tabular}{llr}
\toprule
\textbf{Mode} & \textbf{Source Corpus} & \textbf{Number of Cases} \\
\midrule
Spoken & PDT-C 2.0 & 505 \\
& Czech Academic Corpus 2.0 (CAC) & 7 \\
\cline{2-3}
& \textbf{Total Spoken} & 512 \\
\midrule
\multirow{3}{*}{Written} & PDT-C 2.0 & 461 \\
& iRozhlas & 40 \\
& Czech Academic Corpus 2.0 (CAC) & 11 \\
\cline{2-3}
& \textbf{Total Written} & 512 \\
\midrule
\multicolumn{2}{l}{\textbf{Grand Total}} & \textbf{1,024} \\
\bottomrule
\end{tabular}
}
\end{table}

We also considered the referential status of nominal groups for all categories, and divided the segments according to the specific and generic reference of the nominal groups and pronouns in coreferential relations (see Table~\ref{tab_cor_refstatus}). The division of the specific/generic reference did not apply for the cases with the local pronominal adverb \textit{tam} ‘there'.

\begin{table}[h]
\caption{Structure of Hlava Cor: specific vs. generic reference}
\label{tab_cor_refstatus}
\resizebox{\columnwidth}{!}{%
\begin{tabular}{lr}
\toprule
\textbf{Reference Category} & \textbf{Number of Cases} \\
\midrule
full NPs + $to$ ‘it' with specific reference & 384 \\
full NPs + $to$ ‘it' with generic reference & 384 \\
$tam$ ‘there' (division is not relevant) & 256 \\
\midrule
\textbf{Total Cases} & \textbf{1,024} \\
\bottomrule
\end{tabular}
}
\end{table}

Another dimension was the division of our data into coreference types where the anaphoric expression is a full nominal group (noun, noun + adjective, etc.), pronominal coreference with \textit{to} ‘it', or local coreference with the pronominal adverb \textit{tam} ‘there' (see Table~\ref{tab_cor_np}). %

\begin{table}[h]
\caption{Structure of Hlava Cor: nominal and pronominal anaphors}
\label{tab_cor_np}
\resizebox{\columnwidth}{!} {
\centering
\begin{tabular}{lr}
\toprule
\textbf{Anaphor Type} & \textbf{Number of Cases} \\
\midrule
full NPs & 512 \\
pronominal coreference with $to$ ‘it' & 256 \\
local coreference with $tam$ ‘there' & 256 \\
\midrule
\textbf{Total Segments} & \textbf{1,024} \\
\bottomrule
\end{tabular}
}
\end{table}

Being interested in exploring multiple readings of coreference, we aimed to extract examples that exhibit ambiguous or otherwise vague instances of coreference, and at the same time to allow for comparison with cases where coreference appears to be easily identifiable.
In an effort to assess such a distinction in potential examples automatically, we have first passed the examples to five different Transformer-based coreference resolution models.\footnote{We trained 14 models on CorefUD 1.2~\citelanguageresource{corefud-1.2} using the coreference resolution system CorPipe~\citep{straka-2024-corpipe}, the winner of the CRAC 2024 shared task~\citep{novak-etal-2024-findings}, and selected 5 achieving the best results on the two Czech CorefUD datasets.} All potential examples were pre-selected to be theoretically ambiguous, i.e., based on morphological properties, at least two possible antecedents were present within a five-sentence context, making disagreement %
possible.
Based on the output of the five models, the potential examples were divided into two groups: (i) examples where all five models agreed on the antecedent, and (ii) examples where at least two models disagreed, with preference for examples with larger number of disagreements.
Final examples for annotation were selected from these two groups and included in Hlava Cor in a 1:2 ratio. The ratio was uneven intentionally, as we assumed that cases of models' disagreement would provide more informative material for analysis. Table~\ref{tab_cor_agr} presents the data distribution according to the models' agreement and disagreement.

\begin{table}[h]
\caption{Structure of Hlava Cor: model agreement vs. model disagreement}
\label{tab_cor_agr}
\centering
\begin{tabular}{lc}
\toprule
\textbf{Coreference Status} & \textbf{Number of Cases} \\
\midrule
model agreement  & 352 \\
model disagreement & 672 \\
\midrule
\textbf{Total Cases} & \textbf{1,024} \\
\bottomrule
\end{tabular}
\end{table}

Table~\ref{tab_cor_detail} shows how various categories and dimensions in Tables~\ref{tab_cor_spoken} to~\ref{tab_cor_agr} relate to each other.
For example, the first line states that there are 128 cases with an NP type of the anaphor in spoken data with generic coreference (in 84 of them the coreference resolvers disagreed, in 44 cases they were in agreement).
The features can be combined to larger groups of equal size, for example 'to' spoken (128 cases) vs. 'to' written (128 cases), or 'tam' (256 cases) vs. 'to' (256 cases), or spoken (512 cases) vs. written (512 cases).

\begin{table}[h]
\caption{Numbers of cases for detailed combinations of features in Hlava Cor}
\label{tab_cor_detail}
\resizebox{\columnwidth}{!}{%
\centering
\begin{tabular}{lccc}
\toprule
\textbf{Feature}	&	\textbf{Models}	&	\textbf{Models}	& \textbf{Total }\\
\textbf{Combination}	&	\textbf{Disagreed}	&	\textbf{Agreed}	& \\
\midrule					
NP spoken GEN	&	84	&	44	& 128\\
NP spoken SPEC	&	84	&	44	& 128	\\
NP written GEN	&	84	&	44	& 128	\\
NP written SPEC	&	84	&	44	& 128	\\
tam spoken	&	84	&	44	& 128	\\
tam written	&	84	&	44	& 128	\\
to spoken GEN	&	42	&	22 & 64 \\
to spoken SPEC	&	42	&	22	 & 64\\
to written GEN	&	42	&	22	 & 64\\
to written SPEC	&	42	&	22	 & 64\\
\midrule					
\textbf{Total}	&	\textbf{672}	&	\textbf{352}  & \textbf{1,024}	\\
\bottomrule
\end{tabular}
}
\end{table}

\subsection{Hlava Cor: Inter-Annotator Agreement}
\label{Hlava CorIAA}
For each category described in Section~\ref{statistics} (and summarized in Tables \ref{tab_cor_spoken}, \ref{tab_cor_refstatus}, \ref{tab_cor_np}, and \ref{tab_cor_agr}) we calculate the inter-annotator agreement of our three human annotators for setting the coreferential antecedent.\footnote{~Exact match was required. In order to avoid typo-related errors, the annotators were asked to copy/paste the segments from the original text; also, the disagreements were manually checked by an expert to find and fix cases of insignificant omissions (for example, a comma at the end of the coppied text, a missing character etc.).} Table~\ref{tab_cor_IAA} presents the IAA results for coreference annotation in Hlava Cor.
Overall, the agreement of all three annotators reached 49\%, while partial agreement (at least two out of three annotators) reached 83\%. 
The average pairwise agreement for Hlava Cor reaches 60.4\%.\footnote{This number is not included in Table~\ref{tab_cor_IAA}.} 

\begin{table}[h]
\caption{Hlava Cor: Inter-Annotator Agreement Overview }
\label{tab_cor_IAA}
\resizebox{\columnwidth}{!}{%
\centering
\begin{tabular}{lcc}
\toprule
\textbf{Category} & \textbf{All 3} & \textbf{At Least 2} \\
\midrule

\multicolumn{3}{l}{\textbf{Text Mode (from Table~\ref{tab_cor_spoken})}} \\
\midrule
written data & 45\% & 81\% \\
spoken data & 53\% & 86\% \\

\midrule
\multicolumn{3}{l}{\textbf{Referential Status (from Table~\ref{tab_cor_refstatus})}} \\
\midrule
generic coreference (GEN) & 45\% & 82\% \\
specific coreference (SPEC) & 48\% & 82\% \\
\midrule

\multicolumn{3}{l}{\textbf{Grammatical Form (from Table~\ref{tab_cor_np})}} \\
\midrule
full NPs & 51\% & 86\% \\
pronominal ($to$ ‘it') & 38\% & 75\% \\
local ($tam$ ‘there') & 57\% & 88\% \\

\midrule

\multicolumn{3}{l}{\textbf{Model Status (from Table~\ref{tab_cor_agr})}} \\
\midrule
model agreement & 66\% & 91\% \\
model disagreement & 39\% & 79\% \\
\bottomrule
\textbf{All data} & \textbf{49\%} & \textbf{83\%} \\
\bottomrule
\end{tabular}
}
\end{table}

The inter-annotator agreement results reflect the inherent complexity of coreference annotation, particularly in naturally occurring data and in cases involving vague or underspecified referential relations. When broken down by referential status, agreement for generic coreference (45\%) and specific coreference (48\%) is relatively similar, suggesting that both types of reference pose comparable challenges for human annotation. This contradicts our initial observations and is possibly related to the lexical composition of generic nominal phrases in Hlava Cor. Regarding grammatical form, the pronominal coreference with \textit{to} ‘it' showed the lowest (38\%) agreement as compared to full nominal groups (51\%) and the local adverbial \textit{tam} ‘there' (57\%), confirming that non-nominal and %
underspecified \textit{to} ‘it' may be more ambiguous. A more detailed analysis of these patterns will be the subject of further research.

Another relevant observation supports our hypothesis that instances where the models disagreed tend to be more difficult for human annotators as well. Human annotation on the model-disagreement subset reached only 39\% inter-annotator agreement (IAA), whereas cases where the models agreed showed considerably higher consistency, with an IAA of 66\%.
Notably, an IAA of 66\% is still relatively low, especially given that the models themselves were in agreement. This fact clearly deserves special attention and can be partially explained by the inherent implicitness and ambiguity of coreference relations in our data. Model agreement in such cases likely reflects a combination of chance and biases induced by the training data.
Overall, these findings suggest that model disagreement may serve as an indicator of interpretative complexity in the data.

Taken from a slightly different perspective, we can also examine how many distinct annotation choices were made for a single case, see  Table~\ref{tab:decision_distribution}. %
 The table presents the absolute numbers of unique coreference choices per context annotated by the three annotators. The specific reasons for inter-annotator disagreement are not analyzed in detail here; however, in general, they often stem from differences in the level of detail or interpretation of the referential scope of the entities.

\begin{table}[h]
\caption{Hlava Cor: Distribution of Coreference Choices by Annotator Agreement (absolute numbers).}
\label{tab:decision_distribution}
\centering
\setlength{\tabcolsep}{3pt} %
\resizebox{\columnwidth}{!}{%
\begin{tabular}{llr}
\toprule
\textbf{Choice Type} & \textbf{Agreement Level} & \textbf{Cases} \\
\midrule
One choice & Full (3/3) & 503 \\
Two different antecedent & Partial (2/3) & 351 \\
Three different antecedents & No agreement (1/3) & 170 \\
\midrule
\multicolumn{2}{l}{\textbf{Total Annotated Choices}} & \textbf{1,024} \\
\bottomrule
\end{tabular}
}
\end{table}

\section{Hlava AD: Human Label Variation in Attribution and Discourse}
\label{Hlava AD}

The corpus contains 512 short Czech text segments with explicit inter-sentential discourse relations, multiply annotated by five annotators. All texts in Hlava AD come from PDT-C~2.0.

Similarly to Hlava Cor, we separately considered the written and spoken modalities.

Based on our previous findings of frequent cases of inter-annotator disagreement on discourse relations (\citealp{biblio:ZiTextStructure2024}), we hypothesized that a higher measure of disagreement in inter-sentential relations might be related to the presence of attributive constructions (sentences with verbs of thinking or saying). The supposed reason is that it may be difficult for recipients to recognize whether the upcoming sentence is related to the governing clause (author's speech) or to the dependent direct or indirect speech. 

To analyze the influence of attributive constructions on interpretive variation, we divided the Hlava AD data into two subsets: one containing attributive constructions with verbs of thinking and saying (including both direct and indirect speech) and a control subset with non-attributive constructions.
Table~\ref{tab2-alt} summarizes the structure of Hlava AD concerning the spoken/written and attribution/non-attribution dimensions.

The text segments were chosen from the original source (PDT-C 2.0) based on the syntactic and discourse structure. We searched for explicit inter-sentential discourse relations with a complex sentence in the role of the left argument. This complex sentence contained at least one dependent clause and its governing verb was either a verb of thinking or saying (attributive constructions) or not (non-attributive constructions). In non-attributive items, we filtered out all the text segments containing verbs of thinking or saying in any other role.

\begin{table}
\caption{Structure of the Hlava AD (number of items)}\label{tab2-alt}
\resizebox{\columnwidth}{!}{%
\begin{tabular}{lrrr}
\toprule
 & \multicolumn{1}{l}{Spoken data} & \multicolumn{1}{l}{Written data} & {\textbf{Total}} \\
\midrule
Attribution &  96 & 125 & \textbf{221}\\
Other &  173 & 118 & \textbf{291}\\
\midrule
\textbf{Total} & \textbf{269} & \textbf{243} & \textbf{512}\\
\bottomrule
\end{tabular}
}
\end{table}

\begin{figure*}[t]
\begin{center}
\caption{Annotators' prompt in the Hlava AD annotation task}
\label{fig:prompt}
\begin{tabular}{p{8.3cm} p{3.2cm} p{2.5cm}}
\toprule
Preceding context & Left sentence & Right sentence \\
\midrule
Španělská sekce prestižní Asociace evropských novinářů pozvala již po šesté různorodé spektrum středoevropských politiků, publicistů a filosofů, aby početným posluchačům letního universitního semináře s názvem Střední Evropa mezi Bruselem a Moskvou přednesli své úvahy o tom, co se uprostřed kontinentu děje. Pozornost od počátku budila polská účast. Vítěz nedávných voleb Aleksander Kwasniewski a premiérka poražené vlády Hanna Suchocká, každý ze svého úhlu, ale stejně přesvědčivě popsali, co se stalo v Polsku. &
Kwasniewski opakovaně \textcolor{red}{\textbf{zdůraznil}}, že z cesty zásadních proměn země \textcolor{red}{\textbf{nelze}} sejít: pravice a levice se budou \textcolor{red}{\textbf{přít}} se středem o tempo změn, \textcolor{blue}{\textbf{ale}} o základní vzorec reforem \textcolor{red}{\textbf{není}} sporu. &
Co \textcolor{darkgreen}{\textbf{však}} je vážné, je nevelký zájem veřejnosti o věci veřejné. \\
\bottomrule
\end{tabular}\vspace{2mm}
\end{center}
[\underline{Preceding context:}
The Spanish section of the prestigious Association of European Journalists has invited for the sixth time a diverse spectrum of Central European politicians, publicists and philosophers to present their reflections on what is happening in the middle of the continent to the numerous audience of the summer university seminar entitled Central Europe between Brussels and Moscow. The Polish participation attracted attention from the beginning. The winner of the recent elections, Aleksander Kwasniewski, and the Prime Minister of the defeated government, Hanna Suchocka, each from their own perspective, but equally convincingly described what happened in Poland.\vspace{1mm}\\
\underline{Left sentence}:
Kwasniewski has repeatedly \textcolor{red}{\textbf{emphasized}} that the country \textcolor{red}{\textbf{cannot}} be diverted from the path of fundamental changes: the right and left will \textcolor{red}{\textbf{argue}} with the center about the pace of change, \textcolor{blue}{\textbf{but}} there \textcolor{red}{\textbf{is}} no dispute about the basic formula of reforms.\vspace{1mm}\\
\underline{Right sentence:}
What is serious, \textcolor{darkgreen}{\textbf{however}}, is the public's lack of interest in public affairs.]
\end{figure*}

\subsection{Hlava AD: Annotation Process}
\label{Hlava ADanotproces}

The annotators were shown pairs of sentences which had been related by an explicit discourse relation in the previous annotation of PDT-C 2.0. The \texttt{Right sentence} column contains a discourse connective (marked green, see Figure~\ref{fig:prompt}). The \texttt{Left sentence} column consists of several clauses; in the attributive group of items, the left sentence contains a governing verb of saying / thinking, whereas in the control group, another semantic type of verb is used instead.
For better understanding of the text, annotators were additionally given the three preceding sentences in the \texttt{Preceding context} column. 

The annotators' initial task was to identify the clause in the left sentence, to which the discourse connective relates the right sentence (they searched for the so-called ``target of the relation''). 

To technically simplify the task for the annotators, we pre-marked the finite verbs (or parts of verbal forms) in clauses in the \texttt{Left sentence} in red, see Figure~\ref{fig:prompt}. Thus, the verbs served as identifiers of the whole clauses; annotators entered the chosen verb into the corresponding column. 

In case of coordination between two verbal forms (clauses), the entire construction could be annotated as the target of the relation. Coordination was identified by the pre-marked (blue) conjunctions or punctuation marks. 

Thus, the annotators could mark red verbal forms or blue conjunctive phrases/marks as targets of the relation expressed by the green discourse connective. Additionally, they had a possibility to use an exclamation mark (!) to indicate that the target was in the preceding or far left context, and they used a question mark (?) to signal that they could not find any target in the given text (e.g., they did not understand the text segment as a whole).

The second annotation task was to mark to what degree the annotator needed the previous context (before the left sentence) to understand the discourse relation expressed by the discourse connective. There were three possible answers: 0 (I did not need the previous context); 1 (I needed the previous context to understand and it helped me); and 2 (I needed the previous context to understand, but it did not help me). 

The third annotation task required annotators to explain their choice of the target of the relation by entering comments in a separate plain text field.%

\subsection{Hlava AD: Annotation Example}

Figure~\ref{fig:prompt} presents an example with an attributive construction in the left sentence (\textit{Kwasniewski \textbf{emphasized} that...}). Annotators interpret this segment in two different ways. Some annotators identify the main clause with the governing verb \textit{emphasized} as the target of the discourse relation expressed by the discourse connective \textit{however}. 
Their comments explain that Kwasniewski emphasized political happenings. On the other hand, the annotators stress that what is truly important is what is going on among ordinary people.
This group of annotators does not see the right sentence as a part of the Kwasniewski's speech. 

Conversely, other annotators mark the clause \textit{that the country \textbf{cannot} be diverted...} as the target. They claim in their comments that Kwasniewski knows about the upcoming changes and is concerned about the public's lack of interest in public affairs. Thus, they understand the right sentence as a part of the Kwasniewski's speech. 

\subsection{Hlava AD: Inter-Annotator Agreement}

We calculate the inter-annotator agreement on setting the target of a discoure relation, expressed by the discourse connective in the right sentence.
We can obtain theoretically 1-5 solutions from 5 annotators; nevertheless, some syntactic structures in the left sentences are simpler, they do not provide up to 5 potential targets of the discourse relation. On the other hand, the disagreement can be increased by the possibility to use the mark "?" (not understandable) and "!" (relation to a distant left context). The average pairwise agreement disregarding the syntactic complexity reached 64.9\%. The results for the IAA regarding the syntactic complexity of the left sentence (and, subsequently, a different number of potential targets) is presented in Table~\ref{tab3-IAA_hlava_AD}.

A more detailed analysis of the IAA on the target of a discourse relation can be found in Zikánová et al. \citeyearpar{biblio:ZiNeGoldData2025}, where the multiple annotation serves as a test dataset for the default golden data. According to the authors, the measure of the IAA in Hlava AD is strongly affected by the syntactic complexity  of the left sentence; however, it is  influenced neither by the presence of the attribution, nor by the text mode (spoken or written).

\begin{table}
\caption{IAA on the target of a discourse relation in Hlava AD (percentage points)\label{tab3-IAA_hlava_AD}}
\resizebox{\columnwidth}{!}{%
\begin{tabular}{lrr}
\toprule
 & \multicolumn{1}{l}{Agreement} & \multicolumn{1}{l}{Disagreement} \\
 & \multicolumn{1}{l}{(1 solution from} & \multicolumn{1}{l}{(2-5 solutions from}\\
 & \multicolumn{1}{l}{5 annotators)} & \multicolumn{1}{l}{5 annotators)}\\
\midrule
In general &  38\% & 62\%\\
\midrule
Simple structures &  48\% & 52\%\\
(2 possible targets) &   &\\
Complex structures &  27\% & 73\%\\
(3-10 pos. targets) &  & \\
\bottomrule
\end{tabular}
}
\end{table}

\section{Observations across Hlava Cor and Hlava AD}
Given that Hlava Cor covers a dataset twice the size of Hlava AD, and that Hlava AD was annotated by five annotators while Hlava Cor involved only three parallel annotations, the results are not fully comparable. However, it is possible to compare the average pairwise agreement on target identification, which neutralizes the difference in the number of annotators. For Hlava AD, the average pairwise agreement reaches 64.9\%, while for Hlava Cor it is 60.4\%.

In both corpora, annotators marked the need for context (see Sections~\ref{Hlava Coranotproces} and~\ref{Hlava ADanotproces}). The analysis of the annotations reveals substantial variation across annotators, while the overall tendencies remain similar in both corpora. What we observe likely reflects, to some extent, the annotators’ degree of confidence in their interpretation of the text, as well as individual differences in annotation style. Some annotators tend to consult a broader preceding context to verify their interpretation, while others consider their immediate reading sufficient without referring to additional context. To a certain degree, this also mirrors the annotator’s thoroughness and attention to detail.

Particular attention should be paid to the annotators’ comments, which, in our view, represent the most interesting part of our annotation. Comments were mandatory in both corpora, and they certainly deserve a separate study beyond the scope of this paper. They capture the annotators’ doubts about their chosen decisions, indicate possible alternative interpretations, and often describe in detail the strategies underlying their choices. Especially intriguing is the overlap between the annotators’ comments and their decisions in other tasks. In some cases, the antecedents were chosen identically (inter-annotator agreement), yet the comments still mentioned alternative options. Conversely, in certain cases of inter-annotator disagreement, the analysis shows that the annotators reasoned in a very similar way but ultimately arrived at different decisions.

\section{Conclusions}
In this paper, we presented and described two datasets created to study human label variation: Hlava Cor \citelanguageresource{hlavacor}, focused on coreference, and Hlava AD \citelanguageresource{hlavaad}, focused on attribution and discourse relations. Hlava Cor explores coreference with respect to 
the reference status of anaphoric expressions (specific or generic), their form  (pronouns, full noun phrases, and anaphoric adverbials), and the extent to which coreference resolution models agree on identifying these relations. Hlava AD pays special attention to attribution and non-attributive constructions. Both corpora include spoken and written data and consider both registers equally.

The introduced corpora offer space for a more detailed analysis of label variation in human annotation. Preliminary observations suggest individual differences between annotators and the presence of personal annotation styles. Of particular interest are the annotators’ comments, which often explain annotation choices, reflect subjective judgments, and point to interpretative ambiguity. These and related aspects will be the focus of our future work based on the data obtained from the corpora.

\section*{Acknowledgements}
The authors gratefully acknowledge support from the Grant Agency of Czech Republic, project 24-11132S, as well as the OP~JAK project CZ.02.01.01/00/23\_020/0008518 of the Ministry of Education, Youth and Sports of the Czech Republic.

The research reported in the present contribution has been using language resources developed, stored and distributed by the LINDAT/CLARIAH-CZ project of the Ministry of Education, Youth and Sports of the Czech Republic (LM2023062).

\newpage
\section{Bibliographical References}\label{sec:reference}\vspace{-0.8cm}

\bibliographystyle{lrec2026-natbib}
\bibliography{lrec2026-example}

\section{Language Resource References}
\label{lr:ref}\vspace{-0.8cm}
\bibliographystylelanguageresource{lrec2026-natbib}
\bibliographylanguageresource{languageresource}

\end{document}

%% file: macros_Jirka.tex
\usepackage{ifthen} %

\usepackage[normalem]{ulem}

\usepackage{lingex}

\setlingexlabelmax{9}

\usepackage{enumitem} %
\setlist[itemize]{topsep=5pt}

\usepackage{todonotes}

\definecolor{antiquebrass}{rgb}{0.8, 0.58, 0.46}
\definecolor{ballblue}{rgb}{0.13, 0.67, 0.8}
\definecolor{bittersweet}{rgb}{1.0, 0.44, 0.37}